\documentclass{article}

\usepackage[final, nonatbib]{neurips_2024}

\usepackage[utf8]{inputenc} % allow utf-8 input
\usepackage[T1]{fontenc}    % use 8-bit T1 fonts
\usepackage{hyperref}       % hyperlinks
\usepackage{url}            % simple URL typesetting
\usepackage{booktabs}       % professional-quality tables
\usepackage{amsfonts}       % blackboard math symbols
\usepackage{nicefrac}       % compact symbols for 1/2, etc.
\usepackage{microtype}      % microtypography
\usepackage{xcolor}         % colors
\usepackage{subcaption}
\usepackage{setspace}
\usepackage{graphicx}
\usepackage{multirow}
\usepackage{amsmath}
\usepackage{adjustbox}
\usepackage{caption}
\usepackage{enumitem}

\setlist[itemize]{leftmargin=*}

\title{Lightweight Frequency Masker for Cross-Domain Few-Shot Semantic Segmentation}

\author{\hspace{15pt}
		Jintao Tong
		\hspace{15pt}
		Yixiong Zou\thanks{Corresponding author. \ \ Code is available at \url{https://github.com/TungChintao/APM}.}
		\hspace{15pt}
		Yuhua Li 
		\hspace{15pt}
		Ruixuan Li
		\hspace{15pt}
		\\
		School of Computer Science and Technology, Huazhong University of Science and Technology
		\\
		{\tt\small \{jintaotong, yixiongz, idcliyuhua, rxli\}@hust.edu.cn}
}

\begin{document}

\maketitle

\begin{abstract}
	Cross-domain few-shot segmentation (CD-FSS) is proposed to first pre-train the model on a large-scale source-domain dataset, and then transfer the model to data-scarce target-domain datasets for pixel-level segmentation. The significant domain gap between the source and target datasets leads to a sharp decline in the performance of existing few-shot segmentation (FSS) methods in cross-domain scenarios. In this work, we discover an intriguing phenomenon: simply filtering different frequency components for target domains can lead to a significant performance improvement, sometimes even as high as 14\% mIoU. Then, we delve into this phenomenon for an interpretation, and find such improvements stem from the reduced inter-channel correlation in feature maps, which benefits CD-FSS with enhanced robustness against domain gaps and larger activated regions for segmentation. 
	Based on this, we propose a lightweight frequency masker, which further reduces channel correlations by an Amplitude-Phase Masker (APM) module and an Adaptive Channel Phase Attention (ACPA) module. 
	Notably, APM introduces only 0.01\% additional parameters but improves the average performance by over 10\%, and ACPA imports only 2.5\% parameters but further improves the performance by over 1.5\%, which 
	significantly surpasses the state-of-the-art CD-FSS methods. 
\end{abstract}

\section{Introduction}

Recent advancements in semantic segmentation have been driven by large-scale annotated datasets and developments in deep neural networks \cite{chen2014semantic,long2015fully,zhao2017pspnet,yuan2020object}. Nevertheless, the requirement for extensive labeled data remains a significant challenge, particularly for dense prediction tasks like semantic segmentation.  Hence, few-shot semantic segmentation (FSS) \cite{OSLSM,PL,MAP} has been proposed to meet this challenge, aiming to produce predictions for the unseen categories with only limited annotated data.
However, these FSS methods perform poorly when confronted with domain shifts, particularly when there is a significant gap between the novel class (target domain) and the base class (source domain). This issue has spurred the development of the cross-domain few-shot semantic segmentation (CD-FSS) task \cite{lei2022cross}. Despite various efforts in CD-FSS, the outcomes remain sub-optimal. 

\begin{figure}[t]
	\centering
	%	x`\setlength{\abovecaptionskip}{0.1cm}
	\includegraphics[scale=0.43]{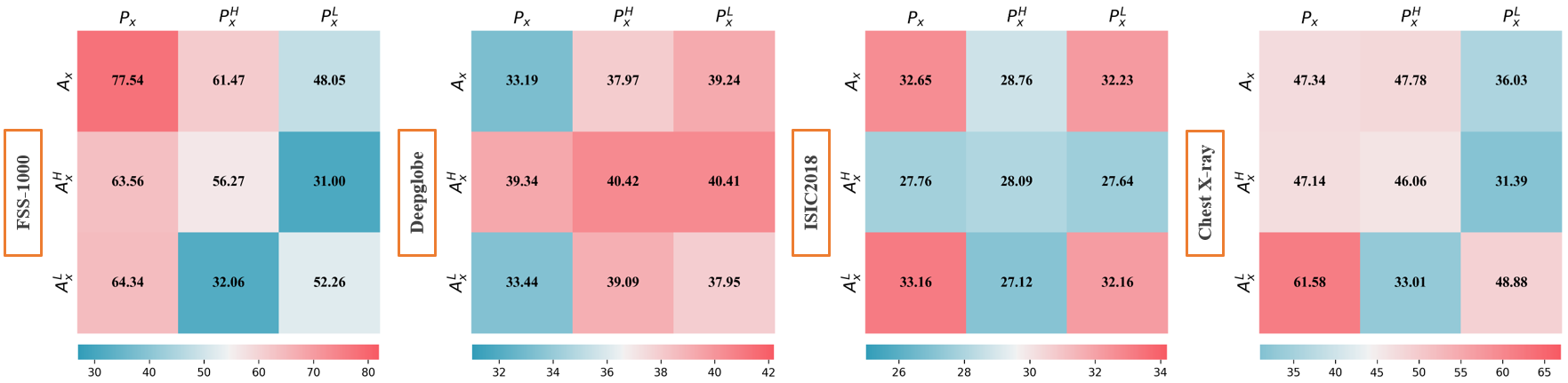}
	\caption{
		For a model already trained on the source domain, we simply filter out different frequency components and plot mIoU against the maintained ones of images. $P$ denotes Phase, $A$ denotes Amplitude, $H$ denotes High Frequency, and $L$ denotes Low Frequency. We can see the performance is significantly improved in most cases compared with the baseline ($A_x$, $P_x$), even as high as 14\% on the Chest X-ray dataset ($A^L_x$, $P_x$). In this paper, we delve into this phenomenon for an interpretation, and propose a lightweight frequency masker for efficient cross-domain few-shot segmentation.
	}
	\label{Fig.finding}
	\vspace{-0.3cm}
\end{figure}

To handle the domain shift problem, efforts have been made to study the generalization of neural networks. Recently, some works~\cite{xu2019frequency,wang2020high,chen2021amplitude} have explored this from the perspective of the frequency domain, achieving theoretical breakthroughs. Compared to humans, neural networks exhibit heightened sensitivity to different frequency components. Additionally, amplitude and phase exhibit distinct properties and effects on neural network performance. 
Inspired by these works, we study the domain shift problem from the perspective of the frequency domain and discover an intriguing phenomenon shown in Figure~\ref{Fig.finding}: for a model already trained on the source domain, \textbf{simply filtering frequency components of images during testing can lead to significant performance improvements}, sometimes even as high as 14\% mIoU.

In this paper, we delve into this phenomenon for an interpretation. Through experiments and mathematical derivations, we find the filtering of the phase and amplitude effectively disentangles feature channels, which lowers the channel correlations and helps the model capture a larger range of semantic patterns.
This benefits the model with improved robustness against large domain gaps, and helps to discover the whole object for segmentation.

Based on the above interpretations, we propose a lightweight frequency masker for the CD-FSS task. This masker does not need to be trained on the source domain, and can be directly inserted into intermediate feature maps during target-domain fine-tuning. It includes an Amplitude-Phase Masker (APM) module and an Adaptive Channel Phase Attention (ACPA) module. 
APM adaptively learns on target domains to filter out harmful amplitude and phase components at a finer granularity, which improves the effectiveness of channel disentanglement.
ACPA learns the attention over channels through phase information.
Notably, the APM module only introduces 0.01\% additional parameters, but can effectively improve the mIoU by over 10\% on average, and ACPA further improves the performance by 1.5\% on average with only 2.5\% additional parameters introduced.

In summary, our contributions can be listed as

\vspace{-0.2cm}
\begin{itemize}[label=$\bullet$]
	\item We find a phenomenon that simply filtering frequency components on target domains can significantly improve performance, with the highest improvement reaching nearly 14\%. 
	
	\item We delve into this phenomenon for an interpretation. We find the frequency filtering operation can effectively disentangle feature map channels, benefiting the model with improved robustness against large domain gaps and a larger range of discovered regions of interest.
	
	\item Based on our interpretations, we propose a lightweight frequency masker for the CD-FSS task, which significantly improves the mIoU by 11\% on average with only 2.5\% parameters introduced.
	
	\item Extensive experiments on four target datasets show that our work, through a simple and effective design, significantly outperforms the state-of-the-art CD-FSS method.
\end{itemize}

\section{Interpreting Enhanced Performance from Frequency Filtering}
In this section, we delve into why filtering certain frequency components can significantly improve CD-FSS performance in certain target domains for interpretation.

\vspace{-0.2cm}
\subsection{Preliminaries}
\vspace{-0.2cm}
Cross-domain few-shot semantic segmentation (CD-FSS) aims to generalize knowledge acquired from source domains with ample training labels to unseen target domains. Given a source domain $ D_s=(\mathcal{X}_s, \mathcal{Y}_s)$ and a target domain $ D_t=(\mathcal{X}_t, \mathcal{Y}_t)$, where $ \mathcal{X} $ represents input data distribution and $ \mathcal{Y} $ represents label space. The model will be trained on the training set from the $ D_s $, then applied to perform segmentation on novel classes in the $ D_t $. Notably, $ D_s $ and $ D_t $ exhibit distinct input data distribution, with their respective label spaces having no intersection, i.e., $ \mathcal{X}_s \neq \mathcal{X}_t $, $ \mathcal{Y}_s \cap \mathcal{Y}_t = \emptyset $.

In this work, we adopt the episodic training manner. Specifically, both the training set sampled from $ D_s $ and the testing set sampled from $ D_t $ are composed of several episodes, each episode is constructed of K support samples $ S=\{I_s^i, M_s^i\}_{i=1}^K $ and a query $ Q=\{I_q, M_q\} $ ($I$ is the image and $M$ is the label).  Within each episode, the model is expected to use $ \{I_s, M_s\} $ and $I_q$ to predict the query label.

\vspace{-0.2cm}
\subsection{Enhanced Performance Stem from Reduced Inter-Channel Correlation}
\vspace{-0.2cm}

Existing research indicates inter-channel relationships are crucial for performance, as different feature channels can represent distinct features~\cite{bau2017network,luo2022channel}. Therefore, we study the change of channel correlations brought by the frequency filtering. 
Specifically, we measured the mean mutual information (MI)~\cite{belghazi2018mutual} between channels from the last layer of the backbone network.
The measured cases include the combination of phase and amplitude for the highest and lowest performance in Fig.~\ref{Fig.finding}.
We report the 1-shot mIoU and MI in Tab.~\ref{tab:mi}, where we can see that for frequency combinations with improved mIoU, their MI consistently decreases, indicating reduced inter-channel correlation in the feature maps. Conversely, for frequency combinations with decreased mIoU, their MI consistently increases. 

The higher the mutual information value, the larger the correlation between channels, while a low MI indicates more independent semantic information captured by different channels. Therefore, the experimental results demonstrate that improved performance is associated with decreased inter-channel correlation in the feature map.

\begin{table}[t]
	\centering
	\caption{Mutual information between feature channels for the best and the worst cases in Fig.~\ref{Fig.finding}. We find that mutual information (MI) consistently decreases when the performance is improved.}
	\label{tab:mi}
	\setstretch{1.2}
	\resizebox{1\linewidth}{!}{
		\begin{tabular}{c|ccc|ccc|ccc|ccc}
			\toprule
			\multirow{2}{*}{Dataset} 
			&\multicolumn{3}{c|}{FSS-1000} 
			&\multicolumn{3}{c|}{Deepglobe} 
			&\multicolumn{3}{c|}{ISIC} 
			&\multicolumn{3}{c}{Chest X-ray} \\
			\cline{2-13}
			&baseline &best &worst
			&baseline &best &worst
			&baseline &best &worst
			&baseline &best &worst\\
			\hline
			1-shot mIoU &77.54&64.34\textcolor{blue}{$\downarrow$}&48.054\textcolor{blue}{$\downarrow$}
			&33.19&40.42\textcolor{red}{$\uparrow$}&33.44\textcolor{red}{$\uparrow$}
			&32.65&33.16\textcolor{red}{$\uparrow$}&27.124\textcolor{blue}{$\downarrow$}
			&47.34&61.58\textcolor{red}{$\uparrow$}&31.394\textcolor{blue}{$\downarrow$} \\
			\hline
			support MI  &1.3736&1.3791\textcolor{red}{$\uparrow$}&1.8767\textcolor{red}{$\uparrow$}  
			&1.3679&1.35024\textcolor{blue}{$\downarrow$} &1.35584\textcolor{blue}{$\downarrow$}
			&1.3789&1.36974\textcolor{blue}{$\downarrow$} &1.3951\textcolor{red}{$\uparrow$} 
			&1.3952&1.39304\textcolor{blue}{$\downarrow$} &1.4315\textcolor{red}{$\uparrow$}\\
			query MI &1.3739&1.3805\textcolor{red}{$\uparrow$} &1.8201\textcolor{red}{$\uparrow$} 
			&1.3667&1.34354\textcolor{blue}{$\downarrow$} &1.35984\textcolor{blue}{$\downarrow$}
			&1.3792&1.36944\textcolor{blue}{$\downarrow$} &1.3890\textcolor{red}{$\uparrow$} 
			&1.3921&1.38774\textcolor{blue}{$\downarrow$} &1.4368\textcolor{red}{$\uparrow$}\\
			\bottomrule
		\end{tabular}
	}
\end{table}
\begin{figure}[t]
	\centering
	%	x`\setlength{\abovecaptionskip}{0.1cm}
	\includegraphics[scale=0.39]{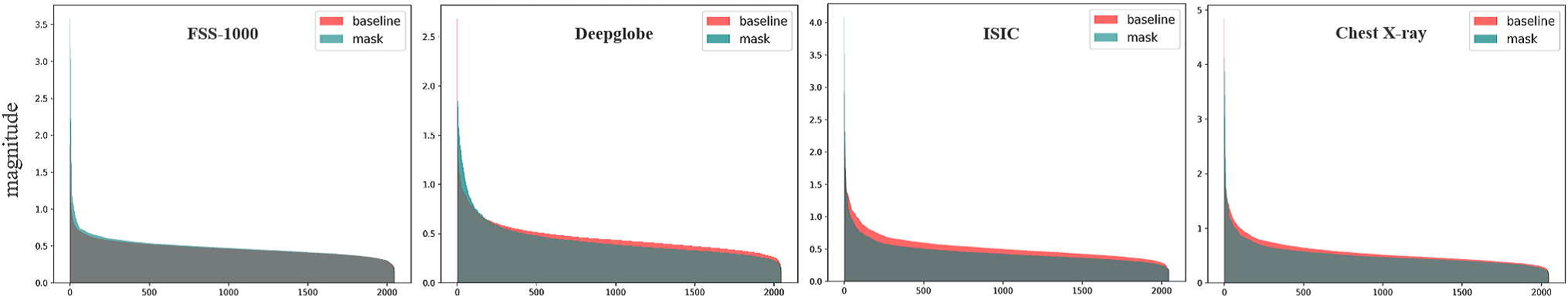}
	%	\vspace{-1em}
	\caption{Mean Magnitude of Channels (MMC) for the best case in Fig.~\ref{Fig.finding} on four target datasets. For domains with improved performance, their curves are lower than the baseline after masking.}
	\label{Fig.mmc}
	\vspace{-0.3cm}
\end{figure}

\vspace{-0.2cm}
\subsection{Why Lower Inter-Channel Correlation is Better?}
\vspace{-0.2cm}

The reduced inter-channel correlation benefits our model in two aspects:

(1) \textbf{Cross-domain generalization}. Previous works~\cite{bardes2021vicreg, zhou2024delve} indicate that a lower correlation between features implies reduced redundancy and enhanced generalizability. 
Intuitively, a lower inter-channel correlation demonstrates that channels capture patterns more independently, therefore each channel will capture patterns in the input image more uniformly, which means the mean magnitude of each channel across all images will be more uniform.
Consistent with our intuition, \cite{luo2022channel} shows the channel bias problem affects the generalizability of few-shot methods, and it utilizes the Mean Magnitude of Channels (MMC) to visualize and measure the channel response in features, where effective few-shot methods might have a more uniform MMC curve in the testing set. 
Therefore, this means the reduced correlation also benefits our model by addressing the channel bias problem, as studied in \cite{luo2022channel}.

Inspired by this, we visualize the MMC before and after applying the mask to filter frequency components on four target datasets. As shown in Figure~\ref{Fig.mmc}, for FSS-1000, performance degrades after masking, with the curve steeper than the baseline. Conversely, for the other three target datasets, performance improves with the curve more uniform than the baseline after masking. This indicates the channel bias problem is also handled by frequency filtering, which benefits the model with more independent and diverse semantic patterns to represent target domains.

(2) \textbf{Exploring larger activated regions for segmentation}.
To study why reducing inter-channel correlation through frequency filtering benefits the segmentation task, we visualize the heatmap of feature maps before and after filtering out specific frequency components. As shown in Figure~\ref{Fig:vis_his}(a), after filtering certain frequency components, the heatmap demonstrates expanded activation regions, which results from more recognition patterns captured by independent channels.
Since the segmentation task requires the model to pixel-wisely detect the whole semantic object, an expanded activation region means the model can better detect the entire object, instead of only focusing on the most discriminative parts.

\vspace{-0.2cm}
\subsection{Why Feature Disentanglement in the Frequency Domain?}
\vspace{-0.2cm}
In this subsection, we illustrate why the feature disentanglement is carried in the frequency domain.
\vspace{-0.2cm}
\paragraph{Fourier Transform (FT)} FT transforms finite signals into complex-valued functions of frequency. For a single channel $ f \in \mathbb{R}^{h \times w}$ of the feature map, the Fourier transform is formulated as:
\begin{equation}
	F(u, v) = \frac{1}{wh} \sum_{x=0}^{w-1} \sum_{y=0}^{h-1} f(x, y) e^{-i 2\pi \left(\frac{ux}{w} + \frac{vy}{h}\right)}
\end{equation}

where $i$ is imaginary unit, $h$ and $w$ are the height and the width of $f$. $f(x,y)$ is an element of $f$ at spatial pixel $(x,y)$, and $F(u,v)$ represents the Fourier coefficient at frequency component $(u,v)$.

The process in which the spatial feature $f$ is decomposed into the amplitude $\alpha$ and phase $\rho$ is called spectral decomposition. The corresponding frequency feature $F$ can be reassembled from amplitude $\alpha$ and phase $\rho$ ($|F|$ is the modulus of $F$):
\begin{equation}
	F = \alpha \cos(\rho) + i \alpha \sin(\rho) = \alpha \cdot e^{i \rho}
\end{equation}
\begin{equation}
	|F| = \sqrt{\alpha^2 (\cos^2(\rho) + \sin^2(\rho))} = \sqrt{\alpha^2} = \alpha
\end{equation}

\textbf{\textit{Mathematical Derivation}}. To intuitively prove the correlation between phase differences and channel correlation within a feature map. For different channels of a feature map, representing distinct features, $F_1(m,n) = \alpha_1 e^{i \rho_1}$ and $F_2(m,n) = \alpha_2 e^{i \rho_2}$ are defined as the frequency domain representations of the same location in different channels. The correlation coefficient is calculated based on the following formula in the frequency domain:
\begin{equation}
	r = \sum_{m=0}^{h-1} \sum_{n=0}^{w-1}\frac{F_1(m,n) F_2^*(m,n)}{\sqrt{|F_1(m,n)|^2 |F_2(m,n)|^2}} = \sum_{m=0}^{h-1} \sum_{n=0}^{w-1} r(m,n)
\end{equation}

The $F_2^*(m,n)$ is the complex conjugate of $F_2(m,n)$, can be computed as:
\begin{equation}
	F_2^*(m,n) = \alpha_2 \cos(\rho_2) - i \alpha_2 \sin(\rho_2) = \alpha_2 e^{-i \rho_2}
\end{equation}

Substituting equations (2), (3), and (5) into equation (4) yields:
\begin{equation}
	F_1(m,n) F_2^*(m,n) = \alpha_1 \alpha_2 e^{i (\rho_1-\rho_2)}
\end{equation}
\begin{equation}
	|F_1(m,n)|^2|F_2(m,n)|^2 = \alpha_1^2 \alpha_2^2
\end{equation}

from which we can further derive:
\begin{equation}
	r(m,n) = \frac{\alpha_1 \alpha_2 e^{i (\rho_1-\rho_2)}}{\sqrt{\alpha_1^2 \alpha_2^2}} = e^{i (\rho_1-\rho_2)}, \ \  \rho_1 - \rho_2 = \Delta\rho \in [0, \pi]
\end{equation}
According to Euler's formula, we know that $e^{\pi i} = -1$ and $e^{0} = 1$. From this derivation, we have proved that the correlation between features in the spatial domain can be translated into phase differences and amplitudes in the frequency domain. When the frequency components are identical, the following can be inferred: 1) $\Delta \rho = 0$, $r(m,n)=1$ indicates perfect positive correlation; 2) $\Delta \rho = \pi$, $r(m,n)=-1$ indicates a perfect negative correlation. Therefore, when the phase differences of more corresponding points from different channels in the frequency domain aggregate around 0 or $\pi$, there is a higher correlation between the channels.
For amplitude, when the phase differences are the same, the closer the amplitudes, the more similar the waveforms are, thus indicating a higher correlation between the channels.

\begin{figure}[t]
	\centering
	\includegraphics[width=\linewidth]{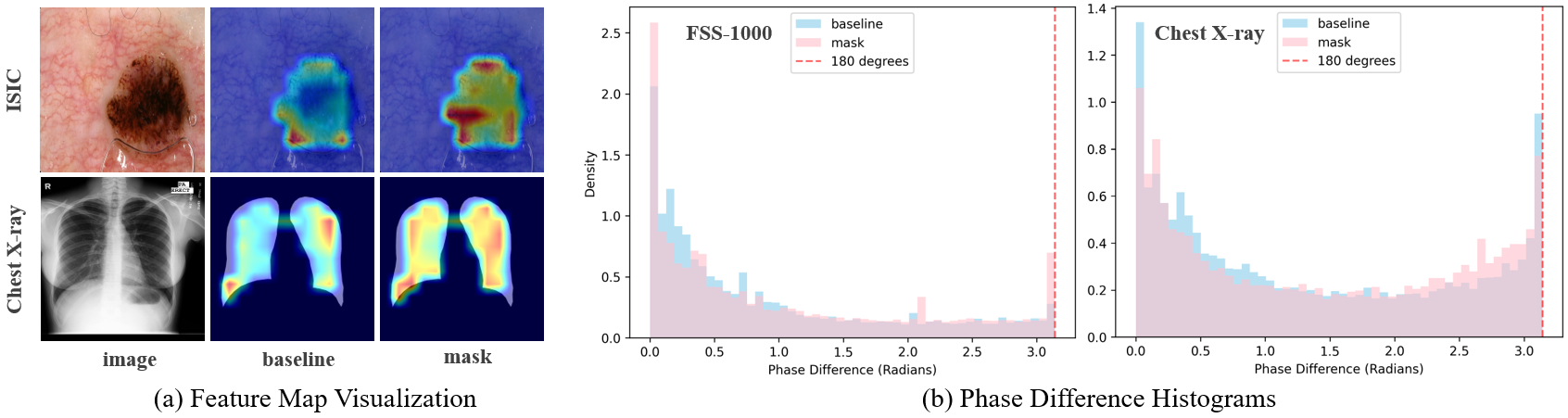}
	\caption{(a) After masking certain frequency components, the model's attention regions are enlarged with more patterns encompassed. (b) A higher concentration of phase differences at 0 and $\pi$ indicates a higher correlation, so that on FSS-1000 the performance drops but on Chest X-ray it increases.}
	\label{Fig:vis_his}
	\vspace{-0.3cm}
\end{figure}

\textbf{Experiments for derivation.} To validate our derivations, we measured the phase difference histograms between channels, using amplitude as weights, to observe the relationship between phase differences among feature map channels. The phase difference and weight are as follows:
\begin{equation}
	\Delta \rho = |\rho_1 - \rho_2|, \ \ weight = \frac{\alpha_1 \alpha_2}{|\alpha_1 - \alpha_2|}
\end{equation}
The results shown in Figure~\ref{Fig:vis_his}(b), indicate that for FSS-1000, after masking certain frequency components, the inter-channel correlation increases. Correspondingly, phase differences between channels tend to cluster more around 0 and $\pi$. Conversely, for Chest X-ray, masking certain frequency components reduces inter-channel correlation, resulting in fewer phase differences clustering around 0 and $\pi$. The experimental results have validated the accuracy of our derived conclusions.

\subsection{Conclusion and Discussion}
\vspace{-0.2cm}

Through mathematical derivation and experiments, we demonstrate that manipulating the frequency can reduce inter-channel correlation in feature maps. Each channel of the feature map represents a distinct pattern, and a lower inter-channel correlation implies a higher degree of channel disentanglement, leading to more independent and diverse semantic patterns for each feature. This benefits the model with 1) alleviation of channel bias, boosting model robustness on target domains; and 2) larger activated regions for segmentation, therefore a simple frequency filtering operation can significantly improve performance for the CD-FSS task.

Based on the above analysis, we can draw the following insights: 1) The aforementioned mask operates at the input level, but fundamentally affects the feature map’s channel correlation. Therefore, we can directly apply mask operations to the frequency domain of each channel in the feature map; 2) Different domains require filtering different components. The aforementioned mask manually filters different frequencies based on the target domain, but the mask can be made adaptive; 3) The aforementioned mask does not perform well on FSS-1000. We believe this may be due to the overly coarse high-low frequency division. A finer frequency division can be designed, dividing the frequency into $h \times w$ parts (where $h$ and $w$ are the spatial dimensions of the feature map).

\section{Method}
\vspace{-0.2cm}

\begin{figure*}[t]
	\centering
	\includegraphics[scale=0.41]{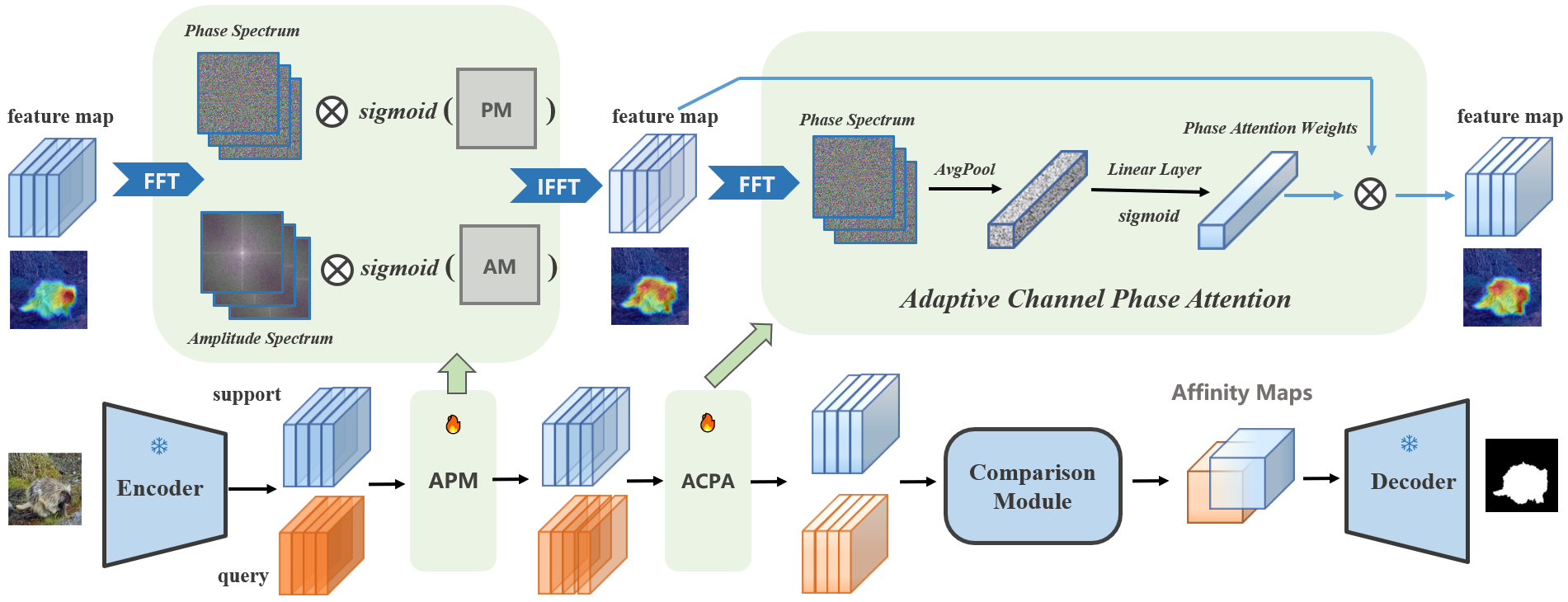}
	%	\vspace{-1em}
	\caption{Overview of our method in a 1-shot example. After obtaining the feature map, APM is introduced to adaptively filter certain frequency components based on different domains, facilitating feature disentanglement to achieve more generalizable representations. Additionally, we propose ACPA to encourage the model to focus on more effective features while aligning the feature space of the support and query images. The internal structure of APM and ACPA is highlighted in green.}
	\label{Fig.overview}
	\vspace{-0.3cm}
\end{figure*}

Our method consists of two major steps, i.e., 1) amplitude-phase masker is proposed to reduce feature correlation, and obtain more accurate and generalized feature maps; 2) adaptive channel phase attention is proposed to select features that benefit the current instance and align the feature spaces of support and query. Our modules do not require source-domain training and can be directly integrated into during target-domain fine-tuning. The overall framework of our approach is shown in Figure \ref{Fig.overview}.
\vspace{-0.2cm}
\subsection{Amplitude-Phase Masker}
\vspace{-0.2cm}
Amplitude-Phase Masker(APM) is a model-agnostic module that filters out negative frequency components at the feature level within feature maps. Through mathematical derivation, it is shown that APM accomplishes feature disentanglement. Consequently, this leads to a feature map that is more robust, generalizable, and provides broader and more accurate representations.

In our work, we utilize a fixed encoder, trained on the source domain, to extract feature maps $\mathcal{F} \in \mathbb{R}^{c \times h \times w}$,  where $c,h$ and $w$ represent channels, height, and width. We then apply the Fast Fourier Transformation (FFT) to convert these feature maps from the spatial domain into the frequency domain, further decomposing them into phase spectrum $\mathcal{P}$ and amplitude spectrum $\mathcal{A}$:
\begin{equation}
	\mathcal{A} e^{i\mathcal{P}} = FFT(\mathcal{F)}
\end{equation}
Then, the phase and amplitude obtained from the FFT are each subjected to a Hadamard product with their respective sigmoid-activated phase mask (PM) $\mathcal{M}_p$ and amplitude mask (AM) $\mathcal{M}_a$. This operation effectively filters out negative components from both the phase and amplitude:
\begin{equation}
	\begin{aligned}
		\mathcal{P}_{enh} &= Sigmoid(\mathcal{M}_p) \otimes \mathcal{P} \\
		\mathcal{A}_{enh} &= Sigmoid(\mathcal{M}_a) \otimes \mathcal{A}
	\end{aligned}
\end{equation}
where $\otimes$ indicates the element-wise multiplication, $\mathcal{P}_{enh}$ and $\mathcal{A}_{enh}$ denotes the enhanced phase and amplitude, respectively. For each task, the APM is initialized with all ones, where $Sigmoid(\mathcal{M}_*) \in [0,1]$. Here, 1 allows complete passage of frequency components, while 0 results in their total filtration. The original APM (APM-S) is a lightweight module, configured as an $h \times w$ matrix that matches the height and width of the feature map. We also offer a variant APM (APM-M) that expands the dimensions to $c \times h \times w$, in alignment with the feature map's dimensions. 

The filtered phase and amplitude components are recombined and transformed back into the spatial domain using the Inverse Fast Fourier Transform (IFFT) to produce the enhanced feature map:
\begin{equation}
	\mathcal{F}_{enh} = IFFT(\mathcal{A}_{enh} e^{i\mathcal{P}_{enh}})
\end{equation}
We iterate and optimize the APM using the support and its corresponding labels. After the APM process, the model generates a feature map that is more accurate and generalizable. This feature map is then fed into the subsequent Adaptive Channel Phase Attention module for further optimization.

\subsection{Adaptive Channel Phase Attention}
\vspace{-0.2cm}
Adaptive Channel Phase Attention(ACPA) can be seen as a process of feature selection. Building on the APM-optimized feature map, ACPA encourages the model to focus on more effective channels (features) while aligning the feature spaces of the support and query. Its underlying insight is that amplitude and phase are considered as style and content, respectively. Therefore, the phase can be seen as an invariant representation, with consistent phase elements across both support and query.

For the enhanced support feature maps $\mathcal{F}_{enh}^{sup}$, a support mask $M^s \in \{0,1\}^{H\times W}$ is applied to discard irrelevant activations:
\begin{equation}
	\mathcal{F}_{mask}^{sup} = \mathcal{F}_{enh}^{sup} \otimes M^s
\end{equation}
Subsequent operations input to ACPA are consistent with those applied to the query feature maps. Therefore, we uniformly refer to the feature map fed into ACPA as $\mathcal{F}_{enh}$. We adopt a SE block following SENet~\cite{hu2018squeeze} as our channel attention module, denoted as $SE$:
\begin{equation}
	\mathcal{W}_{phase} = SE(\mathcal{P}_{enh})
\end{equation}
where $\mathcal{W}_{phase} \in \mathbb{R}^{c \times 1 \times 1}$ is phase attention weight, $\mathcal{P}_{enh}$ is the phase of $\mathcal{F}_{enh}$.
Then we apply the phase attention weights to the feature map $\mathcal{F}_{enh}$ to obtain the final feature map $\mathcal{F}_{final}$:
\begin{equation}
	\mathcal{F}_{final} = \zeta_l(\mathcal{W}_{phase}) \otimes \mathcal{F}_{enh}
\end{equation}
where $\otimes$ is the Hardmard product and $\zeta_l(*)$ extends the weight to match the dimension of the feature map by expanding along the spatial dimension, i.e., $\zeta_l : \mathbb{R}^{c\times 1 \times 1} \rightarrow \mathbb{R}^{c \times h\times w}$.

Finally, a pair of query feature maps $\mathcal{F}_{final}^{qry}$ and support feature maps $\mathcal{F}_{final}^{sup}$ are fed into comparison module forms affinity maps $ C \in \mathbb{R} ^{h \times w \times h \times w} $ using cosine similarity:
\begin{equation}
	C(m,n) = ReLU(\frac{\mathcal{F}_{final}^{qry}(m)\cdot \mathcal{F}_{final}^{sup}(n)}{\lVert \mathcal{F}_{final}^{qry}(m) \rVert \lVert \mathcal{F}_{final}^{sup}(n) \rVert}) 
\end{equation}
where $m,n$ denote 2D spatial positions of feature maps $\mathcal{F}_{final}^{qry}$ and $\mathcal{F}_{final}^{sup}$ respectively. Then, the $ C(m,n) $ is fed into the decoder to obtain segmentation results, as shown in Figure \ref{Fig.overview}.

	\begin{table}[t]
	\centering
	\caption{Mean-IoU of 1-shot and 5-shot results on the CD-FSS benchmark. The best and second-best results are in bold and underlined, respectively. * denotes the model implemented by ourselves. APM-S is an $1 \times h \times w$ matrix, while APM-M (more parameters) expands to $c \times h \times w$.}
	\label{tab:1}
	\setstretch{1.2} 
	\vspace{0.1cm}
	\resizebox{0.9\linewidth}{!}{
		\begin{tabular}{c|cc|cc|cc|cc|cc}
			\toprule
			\multirow{2}{*}{Method}
			&\multicolumn{2}{c|}{FSS-1000} &\multicolumn{2}{|c}{Deepglobe} &\multicolumn{2}{|c}{ISIC} &\multicolumn{2}{|c}{Chest X-ray} &\multicolumn{2}{|c}{Average} \\
			\cline{2-11}
			&1-shot&5-shot &1-shot&5-shot &1-shot&5-shot &1-shot&5-shot &1-shot&5-shot \\
			\hline
			PGNet \cite{pgnet}  &62.42&62.74 &10.73&12.36 &21.86&21.25 &33.95&27.96 &32.24&31.08 \\
			PANet \cite{panet} &69.15&71.68 &36.55&\textbf{45.43} &25.29&33.99 &57.75&69.31 &47.19&55.10 \\
			CaNet \cite{canet} &70.67&72.03 &22.32&23.07 &25.16&28.22 &28.35&28.62 &36.63&37.99 \\
			RPMMs \cite{rpmms} &65.12&67.06 &12.99&13.47 &18.02&20.04 &30.11&30.82 &31.56&32.85 \\
			PFENet \cite{pfenet} &70.87&70.52 &16.88&18.01 &23.50&23.83 &27.22&27.57 &34.62&34.98 \\
			RePRI \cite{repri} &70.96&74.23 &25.03&27.41 &23.27&26.23 &65.08&65.48 &46.09&48.34 \\
			HSNet \cite{min2021hypercorrelation} &77.53&80.99 &29.65&35.08 &31.20&35.10 &51.88&54.36 &47.57&51.38\\
			HSNet$^{*}$ \cite{min2021hypercorrelation} &77.54&80.21 &33.19&36.46 &32.65&35.09 &47.34&48.63 &47.68&50.10\\
			PATNet \cite{lei2022cross}  &\underline{78.59}&\underline{81.23} &37.89&42.97 &41.16&\textbf{53.58} &66.61&70.20 &56.06&61.99 \\
			%			RestNet \cite{huang2023restnet} &Res-50 &\textbf{81.53}&\underline{84.89} &-&- &\textbf{42.25}&51.10 &70.43&73.69 &-&- \\
			\hline
			\textbf{Ours (APM-S)} &78.25&80.29   &\underline{40.77}&44.85   &\underline{41.48}&49.39
			&\underline{75.22}&\underline{76.89} 
			&\underline{58.93}&\underline{62.86}\\
			\textbf{Ours (APM-M)} &\textbf{79.29}&\textbf{81.83}   &\textbf{40.86}&\underline{44.92}   &\textbf{41.71}&\underline{51.16} &\textbf{78.25}&\textbf{82.81} &\textbf{60.03}&\textbf{65.18}\\
			\bottomrule
		\end{tabular}
	}
	\vspace{-0.2cm}
\end{table}

\vspace{-0.1cm}
\section{Experiments}
\vspace{-0.2cm}
\subsection{Datasets}
\vspace{-0.2cm}
We utilize the benchmark established by PATNet \cite{lei2022cross} and adopt the same data preprocessing methods. For training, our source domain is the PASCAL-$5^i$ dataset \cite{OSLSM}, an extended version of PASCAL VOC 2012 \cite{everingham2010pascal} enhanced with additional annotations from the SDS dataset. For evaluation, our target domains include FSS-1000 \cite{li2020fss}, Deepglobe \cite{demir2018deepglobe}, ISIC2018 \cite{codella2019skin,tschandl2018ham10000}, and the Chest X-ray datasets \cite{candemir2013lung,jaeger2013automatic}. See Appendix~\ref{dataset} for more details about datasets.
\vspace{-0.2cm}
\subsection{Implementation Details}
\vspace{-0.2cm}
We employ ResNet-50 \cite{he2016deep} as our encoder, initialized with weights pre-trained on ImageNet \cite{russakovsky2015imagenet}. The training manner is consistent with our baseline model HSNet \cite{min2021hypercorrelation}. To optimize memory usage and speed up training, the spatial sizes of both support and query images are set to 400 × 400. The model is trained using the Adam~\cite{kingma2014adam} optimizer with a learning rate of 1e-3. 

During the adaptation stage, the model initially predicts the support mask and then uses the corresponding label to optimize the APM and ACPA through CE loss. The adaption stage of the APM and the ACPA leverage feature maps from conv5\_x whose channel dimensions are 2048 and spatial size is 13$\times$13, is performed using the Adam optimizer, with learning rates set at 0.1 for Chest X-ray, 0.01 for FSS-1000 and ISIC,  and 1e-5 for Deepglobe. Each task undergoes a total of 60 iterations.
\vspace{-0.2cm}
\subsection{Comparison with State-of-the-Art Works}
\begin{figure}[t]
	\begin{minipage}[t]{0.5\linewidth}
		\vspace{0pt}
		\centering
		\includegraphics[width=\textwidth]{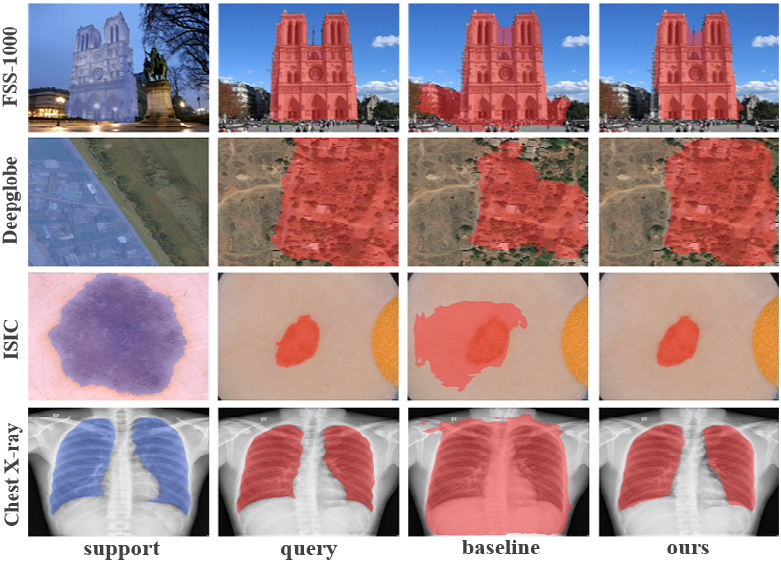}
		\vspace{-0.5cm}
		\caption{Qualitative results of our model.}
		\label{Fig.vis}
	\end{minipage}
	\hfil
	% 第一个表格
	\begin{minipage}[t]{0.45\linewidth}
		\centering
		\captionof{table}{Ablation study on various designs}
		\vspace{-0.1cm}
		\label{tab.ablation}
		\resizebox{0.92\linewidth}{!}{
			\begin{tabular}{ccccc}
				\toprule
				APM-S     &APM-M    & ACPA & 1-shot & 5-shot  \\
				\midrule
				%\hline
				&                &          &47.68 &50.10 \\
				$\checkmark$     &  &       &57.77 &61.39 \\
				$\checkmark$&&$\checkmark$  
				&\textbf{58.93} &\textbf{62.86}\\
				\midrule
				& $\checkmark$    &          &59.13 &63.53 \\ 
				&$\checkmark$&$\checkmark$   &\textbf{60.03}&\textbf{65.18}
				\\
				\bottomrule
			\end{tabular}
		}
		\vspace{0.3cm}  
		\centering
		\caption{APM-S implemented in the transformer architecture.}
		\vspace{-0.1cm}
		\label{tab.vit}
		\resizebox{\linewidth}{!}{
			\begin{tabular}{c|ccccc}
				\toprule
				&FSS &Deepglobe &ISIC &Chest &Average \\
				\midrule
				%\hline
				perSAM~\cite{zhang2023personalize} &79.65 &33.39  &21.27 &31.12 &41.35  \\
				FPTrans~\cite{zhang2022feature} &78.90 &38.29  &47.60 &78.92 &60.93  \\
				FPTrans+ours&\textbf{79.84}&\textbf{39.78} &\textbf{50.37} &\textbf{79.29} &\textbf{62.32}\\
				\bottomrule
		\end{tabular}}
	\end{minipage}
	\vspace{-0.4cm}
\end{figure}

\vspace{-0.2cm}
In Table \ref{tab:1}, we compare our method with several state-of-the-art few-shot semantic segmentation approaches on the benchmark introduced by PATNet \cite{lei2022cross}. Our results show a significant improvement in cross-domain semantic segmentation for both 1-shot and 5-shot tasks. Specifically, APM-S exceeds the performance of the state-of-the-art PATNet, based on ResNet-50, by 2.87\% and 0.87\% in average mIoU for the 1-shot and 5-shot settings, respectively. APM-M outperforms the state-of-the-art by 3.97\% and 3.19\%. Additionally, Figure \ref{Fig.vis} presents qualitative results of our method in 1-way 1-shot segmentation, highlighting a substantial enhancement in generalization across large domain gaps while maintaining comparable accuracy with similar domain shifts.

\subsection{Ablation Study}
\vspace{-0.1cm}
\paragraph{Effectiveness of each module.} 
We evaluated each proposed module in both 1-shot and 5-shot settings to assess their effectiveness. As detailed in Table \ref{tab.ablation}, introducing APM-S increased the average mIoU by 10.09\% for 1-shot and 11.29\% for 5-shot. Adding ACPA further enhanced the mIoU by 1.16\% and 1.47\%, respectively. Additionally, APM-M, a variant of APM with more parameters, when combined with ACPA, increased the average mIoU by 12.35\% for 1-shot and 15.08\% for 5-shot.
\vspace{-0.3cm}
\paragraph{Model-agnostic method.}
We implemented our method for FPTrans~\cite{zhang2022feature}, which employs the Vision Transformer (ViT)~\cite{dosovitskiy2020image} as its encoder. As shown in Table~\ref{tab.vit}, our approach can also effectively enhance the performance of models based on the transformer architecture. Moreover, the new large-scale SAM~\cite{kirillov2023segment} model has significantly advanced image segmentation, showcasing impressive zero-shot capabilities. However, SAM is not suited for cross-domain few-shot segmentation. Thus, we evaluate PerSAM~\cite{zhang2023personalize} to compare our method with the SAM-based approach. The result shows that our method performs much better than PerSAM in cross-domain few-shot segmentation.
\vspace{-0.3cm}
\paragraph{Comparison with other method.}
To demonstrate the effectiveness of our method, we compared it with full parameter fine-tuning and feature disentangling method, as shown in Table~\ref{tab:ft}. Compared to full parameter fine-tuning, our method uses fewer parameters and achieves better performance. For spatial domain feature disentangling, we added a mutual information loss to the baseline model during training to encourage each channel of the feature map to learn independent representations. Our method significantly outperforms this approach. Intuitively, feature disentanglement in the frequency domain offers finer granularity and global representation, which is more beneficial for segmentation tasks compared to the local representation in the spatial domain.

\vspace{-0.2cm}
\subsection{APM: Feature Disentanglement via Frequency Operations}
\begin{figure}[tbp]
	\centering
	%	x`\setlength{\abovecaptionskip}{0.1cm}
	\includegraphics[width=\linewidth]{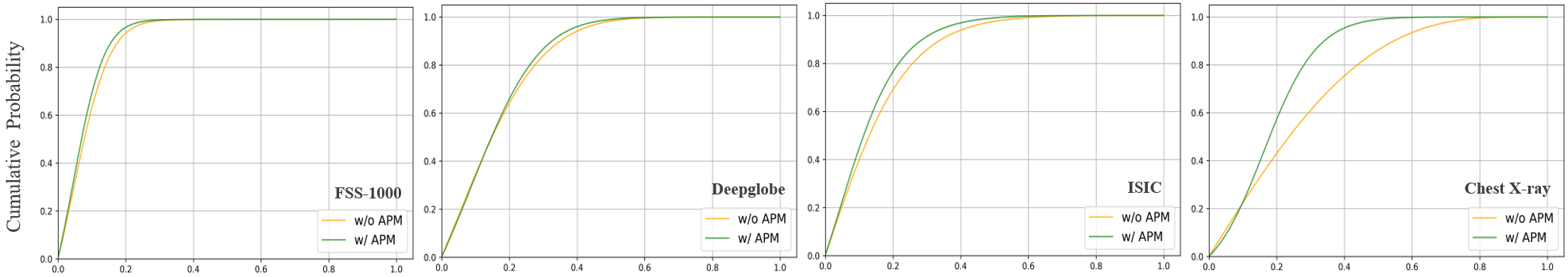}
	%	\vspace{-1em}
	\caption{Cumulative distribution function (CDF) of inter-channel correlations. After passing through APM, the CDF curve shifts to the left, indicating a decrease in inter-channel correlations.}
	\label{Fig.cdf}
	\vspace{-0.2cm}
\end{figure}

\vspace{-0.1cm}
\paragraph{Reduce inter-channel correlation.}
As shown in Table~\ref{tab:mi_apm}, we validated APM's ability to reduce inter-feature correlation and improve generalization performance by calculating the mutual information between feature map channels. Compared to Table~\ref{tab:mi}, this result demonstrates that the adaptive feature-level approach is more effective than the input-level masking, further reducing inter-feature correlation. Furthermore, we plotted the cumulative distribution function (CDF) of inter-channel correlations in the feature maps, as shown in Figure~\ref{Fig.cdf}. It can be observed that with the inclusion of APM, the CDF curve shifts to the left, indicating a decrease in inter-channel correlations.
\vspace{-0.3cm}
\begin{table}[htbp]
	\centering
	\begin{minipage}[t]{0.48\linewidth}
		\centering
		\caption{Compare our method with fine-tuning and spatial domain feature disentangle method.}
		\label{tab:ft}
		\setstretch{1.1}
		\resizebox{\linewidth}{!}{
			\begin{tabular}{c|ccccc}
				\toprule
				&FSS &Deepglobe &ISIC &Chest &Average \\
				\midrule
				%\hline
				baseline &77.54 &33.19  &32.65 &47.34 &47.68  \\
				baseline + ft &78.06 &33.54 &33.28&74.16&54.76\\
				baseline+MI Loss &78.90 &32.28 &32.53 &52.12 &48.76  \\
				ours(APM-M)&\textbf{79.29}&\textbf{40.86} &\textbf{41.71} &\textbf{78.25} &\textbf{60.03}\\
				\bottomrule
			\end{tabular}
		}
	\end{minipage}
	\hfill 
	\begin{minipage}[t]{0.48\linewidth}
		\centering
		\caption{Verify the effectiveness of the APM by the mean mutual information (MI). (w/o ACPA)}
		\label{tab:mi_apm}
		\setstretch{1.25}
		\resizebox{\linewidth}{!}{
			\begin{tabular}{c|cccc}
				\toprule
				APM-S 
				&FSS
				&Deepglobe
				&ISIC
				&Chest \\
				\hline
				1-shot mIoU &77.98$\uparrow$&40.74$\uparrow$&38.79$\uparrow$ &73.55$\uparrow$\\
				support MI  &1.3501$\downarrow$  
				&1.2761$\downarrow$ 
				&1.3139$\downarrow$  &1.2610$\downarrow$\\
				query MI &1.3509$\downarrow$  
				&1.2794$\downarrow$ 
				&1.3121$\downarrow$  &1.2620$\downarrow$\\
				\bottomrule
			\end{tabular}
		}
	\end{minipage}
\end{table}
\paragraph{More independent semantic representations. }
As shown in Table~\ref{tab.heat_cka} (Left), we visualized the heatmaps of feature maps processed by APM and those not processed by APM. It is evident that after applying APM, the model focuses on more different features. For example, in the first column, the baseline only highlights the swan's head, whereas APM makes each feature representation more independent, allowing the model to attend to various features of the swan.

\vspace{-0.1cm}
\subsection{ACPA: Aligning Task-Relevant Features and Feature Spaces}
ACPA can be seen as feature selection, enabling the model to focus on features that are more effective for the current task while aligning the feature spaces of the support and query feature maps. As shown in Table~\ref{tab.heat_cka} (Left), after APM disentangles the features and produces a more broadly represented feature map, ACPA selects features that are more effective and discriminative for the current task. For example, in the first row, ACPA selects the swan's wings and head. In the second column, it selects the bird's head, tail, and feet. Furthermore, we measured the CKA (Centered Kernel Alignment) to calculate the distance between the support feature map and the query feature map, validating that ACPA aligns the support and query feature spaces. CKA is proposed to measure both intra-domain and inter-domain distances~\cite{zou2022margin}; the smaller the CKA value, the closer the feature spaces. As shown in Table~\ref{tab.heat_cka} (Right), after applying APM, the distance between support and query is reduced, and with the addition of ACPA, the support and query feature spaces are further aligned.

\vspace{-0.1cm}
\begin{table}[htbp]
	\begin{minipage}{0.38\linewidth}
		%\vspace{0pt}
		\centering
		\includegraphics[width=\textwidth]{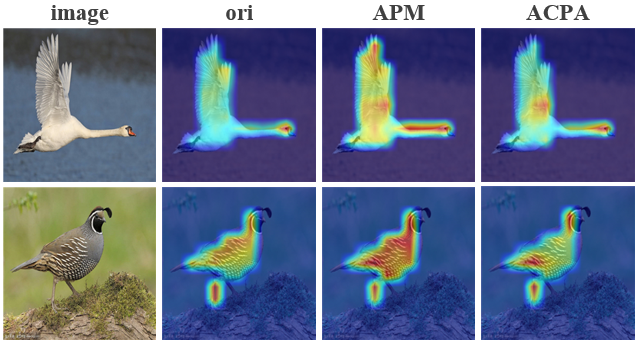}
	\end{minipage}
	\hfill
	\begin{minipage}{0.56\linewidth}
		%\vspace{0pt}
		\centering
		\resizebox{1\linewidth}{!}{
			\begin{tabular}{ccc|cccc}
				\toprule
				APM-S &APM-M    & ACPA &FSS &Deepglobe &ISIC &Chest  \\
				\midrule
				%\hline
				&&               &0.3591 &0.2691&0.2494&0.5848            \\
				$\checkmark$&&&0.3481&0.2678&0.2433&0.5025\\
				$\checkmark$&&$\checkmark$&0.2907&0.2676&0.2032&0.3628\\
				\midrule
				&$\checkmark$ &  &0.3293 &0.2675&0.2310&0.4635       \\
				&$\checkmark$&$\checkmark$&\textbf{0.2883}&\textbf{0.2674} &\textbf{0.1811}&\textbf{0.3986}  \\
				\bottomrule
			\end{tabular}
		}
	\end{minipage}
	\vspace{0.3cm}
	\caption{(Left) Feature map visualizations show: 1) APM achieves feature disentanglement. 2) ACPA encourages the model to focus on more effective features. (Right) Verify the effectiveness of the ACPA in aligning the support and query feature spaces by CKA measure. }
	\label{tab.heat_cka}
	
\end{table}

\vspace{-0.3cm}
\subsection{Comparison with Domain Transfer Methods}
We compare our method against traditional frequency-based and correlation-based approaches to validate our method’s effectiveness. For a fair comparison, all methods are implemented on the baseline HSNet~\cite{min2021hypercorrelation} and then evaluated under the 1-shot setting on the CD-FSS benchmark.

\paragraph{Frequency-based method}
DFF~\cite{lin2023deep} preserves frequency information beneficial for generalization. GFNet~\cite{rao2021global} replaces self-attention layer with global frequency filter layer. ARP~\cite{chen2021amplitude} introduces Amplitude-Phase Recombination via amplitude transformation. DAC~\cite{lee2023decompose} proposes a normalization method that removes style (amplitude) while preserving content (phase) through spectral decomposition. Although these methods improve generalization, they fall short in addressing large domain gaps.
Our method requires no source domain training. It adaptively masks harmful components for the target domain at the feature level. By treating amplitude and phase separately, we exploit phase invariance to design a channel attention module that handles intra-class variations. As shown in Table~\ref{tab:freq} our method outperforms existing frequency-based approaches on the CD-FSS task.
\paragraph{Correlation-based method}
For methods that directly constrain the model (orthogonality, whitening): the few-shot setting means limited sample size, and existing models have a large number of parameters. Directly adjusting the model with constraints using such small datasets is not effective and even can lead to negative optimization. As seen in Table~\ref{tab:reduce}, the performance of orthogonality constraints(SRIP~~\cite{bansal1810can}) and whitening is not satisfactory.
For feature transformation/augmentation methods like MMC~\cite{luo2022channel}: the stability is not guaranteed because they use specific feature transformation functions. Due to the domain gap, a transformation method effective for one domain may not be effective for others. 
In contrast, our method has the advantages of being 1) lightweight (allowing for quick adaptation in the few-shot setting) and 2) stable and robust (with adaptive adjustments for different target domains). These benefits are well reflected in the performance results.

	\begin{table}[t]
	\centering
	\begin{minipage}[t]{0.48\linewidth}
		\centering
		\caption{Compare our method to previous frequency-based methods under 1-shot setting.}
		\label{tab:freq}
		\setstretch{1.1}
		\resizebox{\linewidth}{!}{
			\begin{tabular}{c|ccccc}
				\toprule
				&FSS &Deepglobe &ISIC &Chest &Average \\
				\midrule
				baseline~\cite{min2021hypercorrelation} &77.54 &33.19  &32.65 &47.34 &47.68  \\
				DFF~\cite{lin2023deep} &78.18 &32.16 &35.71&60.29&51.59\\
				GFNet~\cite{rao2021global} &76.86 &32.23 &33.95 &53.12 &49.04  \\
				ARP~\cite{chen2021amplitude} &78.83 &35.06 &35.61 &59.83 &52.33  \\
				DAC~\cite{lee2023decompose} &78.83 &35.98 &36.02 &57.66 &51.98  \\
				ours(APM-M)&\textbf{79.29}&\textbf{40.86} &\textbf{41.71} &\textbf{78.25} &\textbf{60.03}\\
				\bottomrule
			\end{tabular}
		}
	\end{minipage}
	\hfill 
	\begin{minipage}[t]{0.48\linewidth}
		\centering
		\caption{Compare our method to other reducing correlation approaches under 1-shot setting.}
		\label{tab:reduce}
		\setstretch{1.2}
		\resizebox{\linewidth}{!}{
			\begin{tabular}{c|ccccc}
				\toprule
				&FSS &Deepglobe &ISIC &Chest &Average \\
				\midrule
				baseline~\cite{min2021hypercorrelation} &77.54 &33.19  &32.65 &47.34 &47.68  \\
				MMC(Simple)~\cite{luo2022channel} &77.48 &34.70 &34.32 &48.74&48.81\\
				MMC(Oracle)~\cite{luo2022channel} &77.45 &35.12 &34.59 &50.27 &49.36\\
				SRIP~\cite{bansal1810can} &78.13 &34.61 &34.05 &50.58 &49.34  \\
				baseline+whitening &77.92 &33.22 &32.98 &50.89 &48.75  \\
				ours(APM-M)&\textbf{79.29}&\textbf{40.86} &\textbf{41.71} &\textbf{78.25} &\textbf{60.03}\\
				\bottomrule
			\end{tabular}
		}
	\end{minipage}
\end{table}

\section{Related Work}
\vspace{-0.1cm}
\paragraph{Few-shot learning} Few-shot learning aims to build robust representations for new concepts with limited annotated examples. Existing approaches typically fall into three categories: metric learning \cite{snell2017prototypical,vinyals2016matching}, optimization-based methods \cite{maml,ravi2016optimization} and graph-based methods \cite{graph2017few,liu2018graph}. Recently, cross-domain few-shot learning has gained attention due to disparities in both data distribution and label space between meta-testing and meta-training stages. BSCD-FSL \cite{bcdfsl} introduces a challenging benchmark for cross-domain few-shot learning, featuring a substantial domain gap between the source and target domains. It covers several target domains with varying similarities to natural images.
\vspace{-0.1cm}
\paragraph{Few-shot semantic segmentation} Few-shot semantic segmentation aims to segment unseen classes in query images with only a few annotated samples. OSLSM \cite{OSLSM} is the first two-branch FSS model. Following this, PL \cite{PL} introduces a prototype learning paradigm utilizing cosine similarity between pixels and prototypes. SG-One \cite{MAP} adopts masked average pooling (MAP) to optimize the extraction of support features.  Recently, many FSS methods have emerged in the research community, such as RPMMs \cite{rpmms}, PFENet \cite{pfenet}, ASGNet \cite{ASGNet}, and HSNet~\cite{min2021hypercorrelation}. HSNet employs efficient 4D convolutions on multi-level feature correlations, serving as the baseline for our work.
However, these methods primarily address segmenting novel classes within the same domain and struggle with generalization across disparate domains due to significant feature distribution disparities. Bridging this substantial domain gap, particularly with limited labeled data, remains a formidable challenge.
\section{Conclusion}
\vspace{-0.1cm}
In this paper, we delve into the phenomenon that filtering specific frequency components based on different domains significantly improves performance, providing an interpretation through experiment and mathematical derivation.
Building on our interpretation, we propose the APM, a feature-level frequency component mask designed to enhance the generalization of feature map representations.
Further, we introduced ACPA. Based on the APM-optimized feature map, the ACPA encourages the model to focus on more effective features while aligning the feature spaces of the support and query.
Experimental results demonstrate the approach's effectiveness in reducing domain gaps.

%	\begin{ack}
\section*{Acknowledgments}
This work is supported by the National Natural Science Foundation of China under grants 62206102, 62436003, 62376103 and 62302184; the Science and Technology Support Program of Hubei Province under grant 2022BAA046; Hubei Science and Technology Talent Service Project under grant 2024DJC078; and Ant Group through CCF-Ant Research Fund.

%	\end{ack}

%\section*{References}
\bibliographystyle{plain}
\bibliography{APM_ref}

\begin{thebibliography}{10}

\bibitem{bansal1810can}
N~Bansal, X~Chen, and Z~Wang.
\newblock Can we gain more from orthogonality regularizations in training deep
  cnns? arxiv 2018.
\newblock {\em arXiv preprint arXiv:1810.09102}.

\bibitem{bardes2021vicreg}
Adrien Bardes, Jean Ponce, and Yann LeCun.
\newblock Vicreg: Variance-invariance-covariance regularization for
  self-supervised learning.
\newblock {\em arXiv preprint arXiv:2105.04906}, 2021.

\bibitem{bau2017network}
David Bau, Bolei Zhou, Aditya Khosla, Aude Oliva, and Antonio Torralba.
\newblock Network dissection: Quantifying interpretability of deep visual
  representations.
\newblock In {\em Proceedings of the IEEE conference on computer vision and
  pattern recognition}, pages 6541--6549, 2017.

\bibitem{belghazi2018mutual}
Mohamed~Ishmael Belghazi, Aristide Baratin, Sai Rajeshwar, Sherjil Ozair,
  Yoshua Bengio, Aaron Courville, and Devon Hjelm.
\newblock Mutual information neural estimation.
\newblock In {\em International conference on machine learning}, pages
  531--540. PMLR, 2018.

\bibitem{repri}
Malik Boudiaf, Hoel Kervadec, Ziko~Imtiaz Masud, Pablo Piantanida, Ismail
  Ben~Ayed, and Jose Dolz.
\newblock Few-shot segmentation without meta-learning: A good transductive
  inference is all you need?
\newblock In {\em Proceedings of the IEEE/CVF conference on computer vision and
  pattern recognition}, pages 13979--13988, 2021.

\bibitem{candemir2013lung}
Sema Candemir, Stefan Jaeger, Kannappan Palaniappan, Jonathan~P Musco, Rahul~K
  Singh, Zhiyun Xue, Alexandros Karargyris, Sameer Antani, George Thoma, and
  Clement~J McDonald.
\newblock Lung segmentation in chest radiographs using anatomical atlases with
  nonrigid registration.
\newblock {\em IEEE transactions on medical imaging}, 33(2):577--590, 2013.

\bibitem{chen2021amplitude}
Guangyao Chen, Peixi Peng, Li~Ma, Jia Li, Lin Du, and Yonghong Tian.
\newblock Amplitude-phase recombination: Rethinking robustness of convolutional
  neural networks in frequency domain.
\newblock In {\em Proceedings of the IEEE/CVF International Conference on
  Computer Vision}, pages 458--467, 2021.

\bibitem{chen2014semantic}
Liang-Chieh Chen, George Papandreou, Iasonas Kokkinos, Kevin Murphy, and Alan~L
  Yuille.
\newblock Semantic image segmentation with deep convolutional nets and fully
  connected crfs.
\newblock {\em arXiv preprint arXiv:1412.7062}, 2014.

\bibitem{codella2019skin}
Noel Codella, Veronica Rotemberg, Philipp Tschandl, M~Emre Celebi, Stephen
  Dusza, David Gutman, Brian Helba, Aadi Kalloo, Konstantinos Liopyris, Michael
  Marchetti, et~al.
\newblock Skin lesion analysis toward melanoma detection 2018: A challenge
  hosted by the international skin imaging collaboration (isic).
\newblock {\em arXiv preprint arXiv:1902.03368}, 2019.

\bibitem{demir2018deepglobe}
Ilke Demir, Krzysztof Koperski, David Lindenbaum, Guan Pang, Jing Huang, Saikat
  Basu, Forest Hughes, Devis Tuia, and Ramesh Raskar.
\newblock Deepglobe 2018: A challenge to parse the earth through satellite
  images.
\newblock In {\em Proceedings of the IEEE Conference on Computer Vision and
  Pattern Recognition Workshops}, pages 172--181, 2018.

\bibitem{PL}
Nanqing Dong and Eric~P Xing.
\newblock Few-shot semantic segmentation with prototype learning.
\newblock In {\em BMVC}, volume~3, 2018.

\bibitem{dosovitskiy2020image}
Alexey Dosovitskiy, Lucas Beyer, Alexander Kolesnikov, Dirk Weissenborn,
  Xiaohua Zhai, Thomas Unterthiner, Mostafa Dehghani, Matthias Minderer, Georg
  Heigold, Sylvain Gelly, et~al.
\newblock An image is worth 16x16 words: Transformers for image recognition at
  scale.
\newblock {\em arXiv preprint arXiv:2010.11929}, 2020.

\bibitem{everingham2010pascal}
Mark Everingham, Luc Van~Gool, Christopher~KI Williams, John Winn, and Andrew
  Zisserman.
\newblock The pascal visual object classes (voc) challenge.
\newblock {\em International journal of computer vision}, 88:303--338, 2010.

\bibitem{maml}
Chelsea Finn, Pieter Abbeel, and Sergey Levine.
\newblock Model-agnostic meta-learning for fast adaptation of deep networks.
\newblock In {\em International conference on machine learning}, pages
  1126--1135. PMLR, 2017.

\bibitem{graph2017few}
Victor Garcia and Joan Bruna.
\newblock Few-shot learning with graph neural networks.
\newblock {\em arXiv preprint arXiv:1711.04043}, 2017.

\bibitem{bcdfsl}
Yunhui Guo, Noel~CF Codella, Leonid Karlinsky, John~R Smith, Tajana Rosing, and
  Rogerio Feris.
\newblock A new benchmark for evaluation of cross-domain few-shot learning.
\newblock {\em arXiv preprint arXiv:1912.07200}, 2019.

\bibitem{hariharan2011semantic}
Bharath Hariharan, Pablo Arbel{\'a}ez, Lubomir Bourdev, Subhransu Maji, and
  Jitendra Malik.
\newblock Semantic contours from inverse detectors.
\newblock In {\em 2011 international conference on computer vision}, pages
  991--998. IEEE, 2011.

\bibitem{he2016deep}
Kaiming He, Xiangyu Zhang, Shaoqing Ren, and Jian Sun.
\newblock Deep residual learning for image recognition.
\newblock In {\em Proceedings of the IEEE conference on computer vision and
  pattern recognition}, pages 770--778, 2016.

\bibitem{hu2018squeeze}
Jie Hu, Li~Shen, and Gang Sun.
\newblock Squeeze-and-excitation networks.
\newblock In {\em Proceedings of the IEEE conference on computer vision and
  pattern recognition}, pages 7132--7141, 2018.

\bibitem{jaeger2013automatic}
Stefan Jaeger, Alexandros Karargyris, Sema Candemir, Les Folio, Jenifer
  Siegelman, Fiona Callaghan, Zhiyun Xue, Kannappan Palaniappan, Rahul~K Singh,
  Sameer Antani, et~al.
\newblock Automatic tuberculosis screening using chest radiographs.
\newblock {\em IEEE transactions on medical imaging}, 33(2):233--245, 2013.

\bibitem{kingma2014adam}
Diederik~P Kingma and Jimmy Ba.
\newblock Adam: A method for stochastic optimization.
\newblock {\em arXiv preprint arXiv:1412.6980}, 2014.

\bibitem{kirillov2023segment}
Alexander Kirillov, Eric Mintun, Nikhila Ravi, Hanzi Mao, Chloe Rolland, Laura
  Gustafson, Tete Xiao, Spencer Whitehead, Alexander~C Berg, Wan-Yen Lo, et~al.
\newblock Segment anything.
\newblock In {\em Proceedings of the IEEE/CVF International Conference on
  Computer Vision}, pages 4015--4026, 2023.

\bibitem{lee2023decompose}
Sangrok Lee, Jongseong Bae, and Ha~Young Kim.
\newblock Decompose, adjust, compose: Effective normalization by playing with
  frequency for domain generalization.
\newblock In {\em Proceedings of the IEEE/CVF conference on computer vision and
  pattern recognition}, pages 11776--11785, 2023.

\bibitem{lei2022cross}
Shuo Lei, Xuchao Zhang, Jianfeng He, Fanglan Chen, Bowen Du, and Chang-Tien Lu.
\newblock Cross-domain few-shot semantic segmentation.
\newblock In {\em European Conference on Computer Vision}, pages 73--90.
  Springer, 2022.

\bibitem{ASGNet}
Gen Li, Varun Jampani, Laura Sevilla-Lara, Deqing Sun, Jonghyun Kim, and
  Joongkyu Kim.
\newblock Adaptive prototype learning and allocation for few-shot segmentation.
\newblock In {\em Proceedings of the IEEE/CVF conference on computer vision and
  pattern recognition}, pages 8334--8343, 2021.

\bibitem{li2020fss}
Xiang Li, Tianhan Wei, Yau~Pun Chen, Yu-Wing Tai, and Chi-Keung Tang.
\newblock Fss-1000: A 1000-class dataset for few-shot segmentation.
\newblock In {\em Proceedings of the IEEE/CVF conference on computer vision and
  pattern recognition}, pages 2869--2878, 2020.

\bibitem{lin2023deep}
Shiqi Lin, Zhizheng Zhang, Zhipeng Huang, Yan Lu, Cuiling Lan, Peng Chu,
  Quanzeng You, Jiang Wang, Zicheng Liu, Amey Parulkar, et~al.
\newblock Deep frequency filtering for domain generalization.
\newblock In {\em Proceedings of the IEEE/CVF conference on computer vision and
  pattern recognition}, pages 11797--11807, 2023.

\bibitem{liu2018graph}
Yanbin Liu, Juho Lee, Minseop Park, Saehoon Kim, Eunho Yang, Sung~Ju Hwang, and
  Yi~Yang.
\newblock Learning to propagate labels: Transductive propagation network for
  few-shot learning.
\newblock {\em arXiv preprint arXiv:1805.10002}, 2018.

\bibitem{long2015fully}
Jonathan Long, Evan Shelhamer, and Trevor Darrell.
\newblock Fully convolutional networks for semantic segmentation.
\newblock In {\em Proceedings of the IEEE conference on computer vision and
  pattern recognition}, pages 3431--3440, 2015.

\bibitem{luo2022channel}
Xu~Luo, Jing Xu, and Zenglin Xu.
\newblock Channel importance matters in few-shot image classification.
\newblock In {\em International conference on machine learning}, pages
  14542--14559. PMLR, 2022.

\bibitem{min2021hypercorrelation}
Juhong Min, Dahyun Kang, and Minsu Cho.
\newblock Hypercorrelation squeeze for few-shot segmentation.
\newblock In {\em Proceedings of the IEEE/CVF international conference on
  computer vision}, pages 6941--6952, 2021.

\bibitem{rao2021global}
Yongming Rao, Wenliang Zhao, Zheng Zhu, Jiwen Lu, and Jie Zhou.
\newblock Global filter networks for image classification.
\newblock {\em Advances in neural information processing systems}, 34:980--993,
  2021.

\bibitem{ravi2016optimization}
Sachin Ravi and Hugo Larochelle.
\newblock Optimization as a model for few-shot learning.
\newblock In {\em International conference on learning representations}, 2016.

\bibitem{russakovsky2015imagenet}
Olga Russakovsky, Jia Deng, Hao Su, Jonathan Krause, Sanjeev Satheesh, Sean Ma,
  Zhiheng Huang, Andrej Karpathy, Aditya Khosla, Michael Bernstein, et~al.
\newblock Imagenet large scale visual recognition challenge.
\newblock {\em International journal of computer vision}, 115:211--252, 2015.

\bibitem{OSLSM}
Amirreza Shaban, Shray Bansal, Zhen Liu, Irfan Essa, and Byron Boots.
\newblock One-shot learning for semantic segmentation.
\newblock {\em arXiv preprint arXiv:1709.03410}, 2017.

\bibitem{snell2017prototypical}
Jake Snell, Kevin Swersky, and Richard Zemel.
\newblock Prototypical networks for few-shot learning.
\newblock {\em Advances in neural information processing systems}, 30, 2017.

\bibitem{pfenet}
Zhuotao Tian, Hengshuang Zhao, Michelle Shu, Zhicheng Yang, Ruiyu Li, and Jiaya
  Jia.
\newblock Prior guided feature enrichment network for few-shot segmentation.
\newblock {\em IEEE transactions on pattern analysis and machine intelligence},
  44(2):1050--1065, 2020.

\bibitem{tschandl2018ham10000}
Philipp Tschandl, Cliff Rosendahl, and Harald Kittler.
\newblock The ham10000 dataset, a large collection of multi-source
  dermatoscopic images of common pigmented skin lesions.
\newblock {\em Scientific data}, 5(1):1--9, 2018.

\bibitem{vinyals2016matching}
Oriol Vinyals, Charles Blundell, Timothy Lillicrap, Daan Wierstra, et~al.
\newblock Matching networks for one shot learning.
\newblock {\em Advances in neural information processing systems}, 29, 2016.

\bibitem{wang2020high}
Haohan Wang, Xindi Wu, Zeyi Huang, and Eric~P Xing.
\newblock High-frequency component helps explain the generalization of
  convolutional neural networks.
\newblock In {\em Proceedings of the IEEE/CVF conference on computer vision and
  pattern recognition}, pages 8684--8694, 2020.

\bibitem{panet}
Kaixin Wang, Jun~Hao Liew, Yingtian Zou, Daquan Zhou, and Jiashi Feng.
\newblock Panet: Few-shot image semantic segmentation with prototype alignment.
\newblock In {\em proceedings of the IEEE/CVF international conference on
  computer vision}, pages 9197--9206, 2019.

\bibitem{xu2019frequency}
Zhi-Qin~John Xu, Yaoyu Zhang, Tao Luo, Yanyang Xiao, and Zheng Ma.
\newblock Frequency principle: Fourier analysis sheds light on deep neural
  networks.
\newblock {\em arXiv preprint arXiv:1901.06523}, 2019.

\bibitem{rpmms}
Boyu Yang, Chang Liu, Bohao Li, Jianbin Jiao, and Qixiang Ye.
\newblock Prototype mixture models for few-shot semantic segmentation.
\newblock In {\em Computer Vision--ECCV 2020: 16th European Conference,
  Glasgow, UK, August 23--28, 2020, Proceedings, Part VIII 16}, pages 763--778.
  Springer, 2020.

\bibitem{yuan2020object}
Yuhui Yuan, Xilin Chen, and Jingdong Wang.
\newblock Object-contextual representations for semantic segmentation.
\newblock In {\em Computer Vision--ECCV 2020: 16th European Conference,
  Glasgow, UK, August 23--28, 2020, Proceedings, Part VI 16}, pages 173--190.
  Springer, 2020.

\bibitem{pgnet}
Chi Zhang, Guosheng Lin, Fayao Liu, Jiushuang Guo, Qingyao Wu, and Rui Yao.
\newblock Pyramid graph networks with connection attentions for region-based
  one-shot semantic segmentation.
\newblock In {\em Proceedings of the IEEE/CVF International Conference on
  Computer Vision}, pages 9587--9595, 2019.

\bibitem{canet}
Chi Zhang, Guosheng Lin, Fayao Liu, Rui Yao, and Chunhua Shen.
\newblock Canet: Class-agnostic segmentation networks with iterative refinement
  and attentive few-shot learning.
\newblock In {\em Proceedings of the IEEE/CVF conference on computer vision and
  pattern recognition}, pages 5217--5226, 2019.

\bibitem{zhang2022feature}
Jian-Wei Zhang, Yifan Sun, Yi~Yang, and Wei Chen.
\newblock Feature-proxy transformer for few-shot segmentation.
\newblock {\em Advances in neural information processing systems},
  35:6575--6588, 2022.

\bibitem{zhang2023personalize}
Renrui Zhang, Zhengkai Jiang, Ziyu Guo, Shilin Yan, Junting Pan, Xianzheng Ma,
  Hao Dong, Peng Gao, and Hongsheng Li.
\newblock Personalize segment anything model with one shot.
\newblock {\em arXiv preprint arXiv:2305.03048}, 2023.

\bibitem{MAP}
Xiaolin Zhang, Yunchao Wei, Yi~Yang, and Thomas~S Huang.
\newblock Sg-one: Similarity guidance network for one-shot semantic
  segmentation.
\newblock {\em IEEE transactions on cybernetics}, 50(9):3855--3865, 2020.

\bibitem{zhao2017pspnet}
Hengshuang Zhao, Jianping Shi, Xiaojuan Qi, Xiaogang Wang, and Jiaya Jia.
\newblock Pyramid scene parsing network.
\newblock In {\em Proceedings of the IEEE conference on computer vision and
  pattern recognition}, pages 2881--2890, 2017.

\bibitem{zhou2024delve}
Haichen Zhou, Yixiong Zou, Ruixuan Li, Yuhua Li, and Kui Xiao.
\newblock Delve into base-novel confusion: Redundancy exploration for few-shot
  class-incremental learning.
\newblock {\em arXiv preprint arXiv:2405.04918}, 2024.

\bibitem{zou2022margin}
Yixiong Zou, Shanghang Zhang, Yuhua Li, and Ruixuan Li.
\newblock Margin-based few-shot class-incremental learning with class-level
  overfitting mitigation.
\newblock {\em Advances in neural information processing systems},
  35:27267--27279, 2022.

\end{thebibliography}

%\medskip
%
%
%{
	%\small
	%
	%}

%%%%%%%%%%%%%%%%%%%%%%%%%%%%%%%%%%%%%%%%%%%%%%%%%%%%%%%%%%%%
\clearpage
\appendix

\section{Appendix / supplemental material}
\subsection{More Dataset Details}\label{dataset}

\begin{figure}[htbp]
	%是可选项 h表示的是here在这里插入，t表示的是在页面的顶部插入
	\centering
	\includegraphics[width=\linewidth]{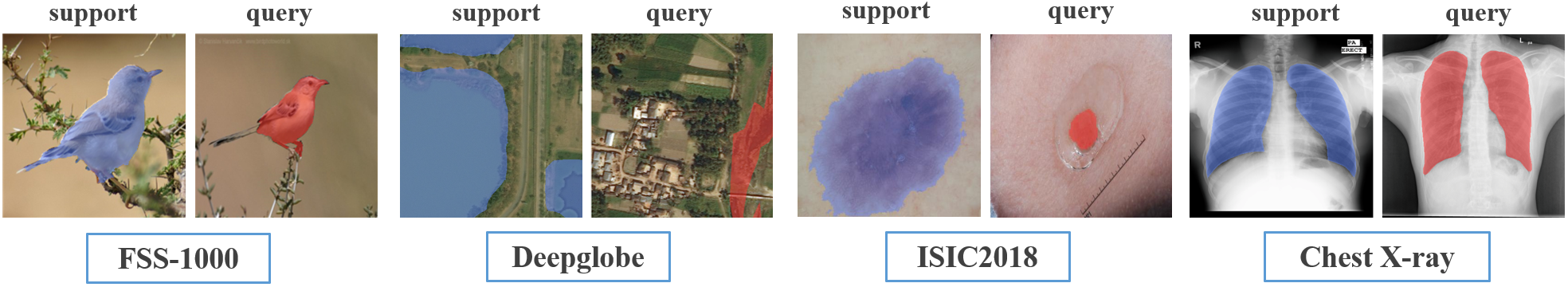}
	\caption{Examples of segmentation for four target datasets.}
	\label{Fig.dataset}
\end{figure}
Our experimental setup is grounded in the benchmark established by PATNet \cite{lei2022cross}. Fig. \ref{Fig.dataset} presents an example of segmentation for four target datasets. Further details follow:

\textbf{PASCAL-$5^i$} \cite{OSLSM} extends the PASCAL VOC 2012 \cite{everingham2010pascal} by integrating additional annotations from the SDS dataset \cite{hariharan2011semantic}. Utilizing PASCAL-$5^i$ as our source domain for training, we then evaluate the models' performance across four target datasets.

\textbf{FSS-1000} \cite{li2020fss} is a few-shot segmentation dataset comprising 1000 natural image categories, with each category containing 10 samples. In our experiment, we adhere to the official split for semantic segmentation and report results on the designated testing set, which encompasses 240 classes and 2,400 images. We consider FSS-1000 as our designated target domain for testing.

\textbf{Deepglobe} \cite{demir2018deepglobe} consists of satellite images annotated densely at the pixel level across 7 categories: urban, agriculture, rangeland, forest, water, barren, and unknown. Since ground-truth labels are available only in the training set, we utilize the official training dataset comprising 803 images to showcase our results. We designate it as our testing target domain and follow the same processing approach as PATNet.

\textbf{ISIC2018} \cite{codella2019skin,tschandl2018ham10000} is designed specifically for skin cancer screening and comprises images of lesions, with each image depicting exactly one primary lesion. Adhering to the guidelines established by PATNet, we process and utilize the dataset, considering ISIC2018 as our target domain for testing.

\textbf{Chest X-ray} \cite{candemir2013lung,jaeger2013automatic} is tailored for Tuberculosis diagnosis, comprising 566 images with a resolution of 4020 × 4892 pixels. These images depict cases from 58 Tuberculosis patients and 80 individuals with normal conditions. A common approach to handling large image sizes involves resizing them to 1024 × 1024 pixels.

\subsection{APM Reduces Inter-Channel Correlation by Frequency Domain Mask}

\begin{figure}[htbp]
	\centering
	\includegraphics[width=\linewidth]{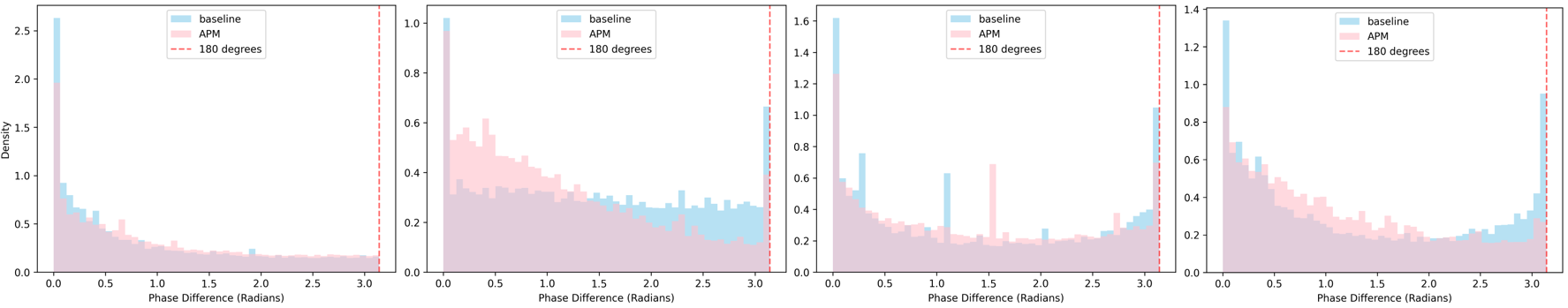}
	\caption{Histogram of phase differences (weighted by amplitude) between channels in the feature maps before and after APM. }
	\label{Fig.appendix_diff}
\end{figure}

As shown in Figure~\ref{Fig.appendix_diff}, we present histograms of the phase differences between channels in the feature maps (weighted by amplitude) before and after adding the APM module. After applying APM, the phase differences between channels concentrated around 0 and $\pi$ are reduced, which aligns with our mathematical derivation in the main text. The APM decreases the correlation between feature map channels by altering phase and amplitude, thereby enhancing the independence of their semantic representations. Additionally, due to the use of finer-grained frequency domain partitioning, APM performs well on FSS-1000 compared to the simple high-low frequency partitioning at the input level.
\subsection{Analyze the Frequency Components Filtered by the APM}
\begin{figure}[t]
	\centering
	\includegraphics[scale=0.38]{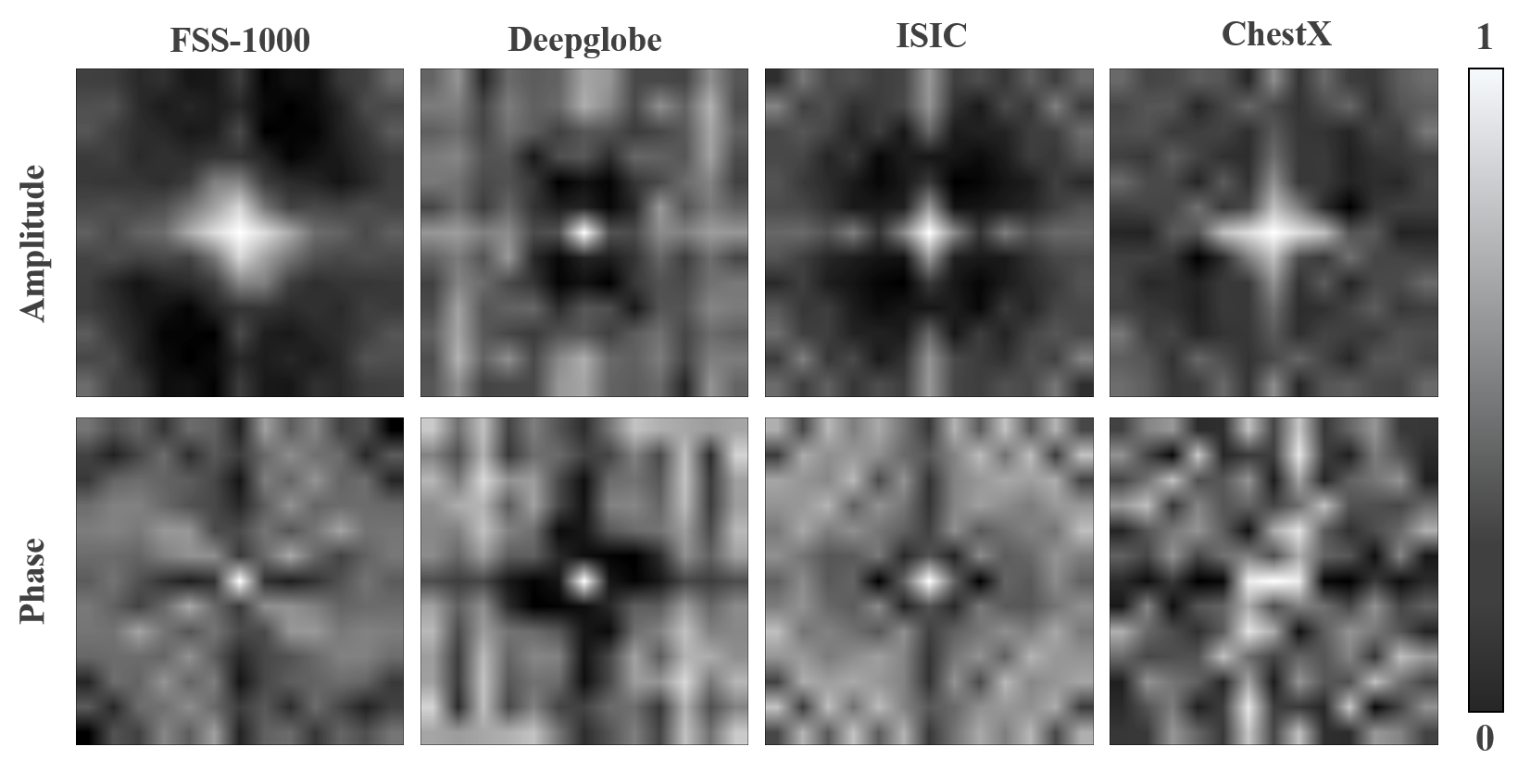}
	\vspace{-0.2cm}
	\caption{The visualization of the frequency components filtered by the masker.}
	\label{Fig.masker}
	\vspace{-0.3cm}
\end{figure}

\vspace{-0.1cm}
The visualization results in Figure~\ref{Fig.masker} show the average frequency components filtered by the amplitude masker and phase masker across different domains. The center represents low frequencies, while the periphery represents high frequencies. White indicates a value of 1, meaning the frequency component passes through, and black indicates a value of 0, meaning the frequency component is filtered out. For FSS, the amplitude masker (AM) primarily filters out mid-to-high-frequency components, while the phase masker (PM) mainly retains mid-frequency components. For DeepGlobe, both AM and PM retain more mid-to-high frequencies. For ISIC, AM filters out more mid-to-high frequencies, retaining low frequencies, whereas PM retains relatively more mid frequencies. For ChestX, AM mainly retains low-frequency components, while PM filters out frequencies across the spectrum, retaining relatively more low-to-mid frequencies. These results align well with the patterns observed in Figure 1 of our main text. It is evident that for different targets, AM and PM dynamically and adaptively filter different frequency components, selecting those more beneficial for the current domain. Additionally, the advantageous amplitude and phase frequency components vary across different target domains, underscoring the necessity of considering amplitude and phase separately.

\vspace{-0.2cm}
\subsection{Detailed Ablation Study Results}
\vspace{-0.1cm}
In the main text, we demonstrate the effectiveness of our various designs by presenting the average mIoU. Here, as shown in Table~\ref{tab:all_dataset}, we provide detailed results on each target dataset.
\vspace{-0.3cm}
\begin{table}[htbp!]
	\centering
	\setstretch{1.2}
	\caption{Detailed ablation study results of various designs (Backbone: ResNet-50). }
	\vspace{0.1cm}
	\label{tab:all_dataset}
	\resizebox{0.85\linewidth}{!}{
		\begin{tabular}{ccc|cc|cc|cc|cc|cc}
			\toprule
			\multirow{2}{*}{APM-S}
			&\multirow{2}{*}{APM-M}      
			&\multirow{2}{*}{ACPA} 
			&\multicolumn{2}{c|}{FSS-1000} &\multicolumn{2}{|c}{Deepglobe} &\multicolumn{2}{|c}{ISIC} &\multicolumn{2}{|c}{Chest X-ray}
			&\multicolumn{2}{|c}{Average}  \\
			\cline{4-13}
			&&&1-shot&5-shot &1-shot&5-shot &1-shot&5-shot &1-shot&5-shot &1-shot&5-shot\\
			\hline			
			&      &          &77.54 &80.21 &33.19&36.46 &32.65 &35.09 &47.34&48.63 &47.68&50.10\\
			$\checkmark$ &      &                    &77.98&79.85 &40.74&44.80 &38.79&44.16 &73.55 &76.73 &57.77&61.39\\
			$\checkmark$ & &$\checkmark$             &78.25&80.29 &40.77&44.85 &41.48&49.39 &75.22&76.89  &58.93&62.86\\
			\hline
			&$\checkmark$ &             &78.98&81.21 &40.81&44.82 &38.99&45.49 &77.73&82.60 &59.13&63.53\\
			&$\checkmark$ &$\checkmark$ &79.29&81.83 &40.86&44.92 &41.71&51.16 &78.25&82.81
			&60.03&65.18\\
			\bottomrule
	\end{tabular}}
\end{table}
\vspace{-0.2cm}
\subsection{Broader Impact}
\vspace{-0.1cm}
Our research delves into the phenomenon that filtering specific frequency components based on different domains significantly improves performance, providing an interpretation. Furthermore, we demonstrated the relationship between frequency components and inter-channel correlation through mathematical derivation. 
Building on our interpretation and derivation, we propose the APM, a feature-level frequency component mask designed to enhance the generalization of feature map representations.
Further, we introduced Adaptive Channel Phase Attention (ACPA). Based on the APM-optimized feature map, the ACPA encourages the model to focus on more effective features while aligning the feature spaces of the support and query.
Experimental results demonstrate the effectiveness of our approach significantly enhances the model’s cross-domain transferability.
This work is applicable not only to CDFSS but also to other areas like domain generalization and domain adaptation. Future research will aim to broaden our evaluations to encompass a wider range of target domains, enhancing our understanding of their performance in various real-world scenarios.

\end{document}